%% file: main.tex
\documentclass{llncs}
\usepackage[T1]{fontenc}
\usepackage{amsmath,amssymb,amsfonts}
\usepackage{graphicx}
\usepackage{textcomp}
\usepackage[table]{xcolor}
\usepackage{booktabs}
\usepackage{multirow}
\usepackage{placeins}
\usepackage{bbding}
\usepackage[hidelinks]{hyperref}
\begin{document}
\pagestyle{empty}

\title{Aerial Content-Aware Sample-Adaptive Classifier for Urban Photogrammetric Point Cloud Segmentation}
\titlerunning{ACASAC}

\author{Chong Liu\inst{1,2,3} \and 
Yonghua Lu\inst{4} \and 
Zhanglin Cheng\inst{3}\thanks{Corresponding author: \email{zl.cheng@siat.ac.cn}}}

\authorrunning{C. Liu et al.}

\institute{
$^{1}$Southern University of Science and Technology\\
$^{2}$Shenzhen University of Advanced Technology\\
$^{3}$Shenzhen VisuCA Key Lab, Shenzhen Institutes of Advanced Technology, \\
Chinese Academy of Sciences\\
$^{4}$Urban Planning \& Design Institute of Shenzhen
}

\maketitle
\thispagestyle{empty}

\input{abstract.tex}

\input{introduction.tex}

\input{related_works.tex}

\input{method.tex}

\input{experiments.tex}

\FloatBarrier
\section{Conclusion}
In this paper, we present the Aerial Content-Aware Sample-Adaptive Classifier (ACASAC) for multimodal urban point-cloud segmentation. ACASAC combines the fixed fused-head prior with scene-level aerial and relation-transferred point-cloud prototypes to adapt the classifier to each paired sample. Experiments on SensatUrban and UrbanBIS, together with prototype and encoder ablations, demonstrate consistent improvements over matched fixed and adaptive heads within the evaluated settings.

\begin{credits}
\subsubsection{\ackname}
This work was supported in part by Guangdong Major Project of Basic and Applied Basic Research (2023B0303000016), and the Shenzhen Science and Technology Program (CJGJZD20240729141906008).
\end{credits}

\newpage

\begingroup
\renewcommand{\doi}[1]{\href{https://doi.org/#1}{\nolinkurl{#1}}}
\linespread{0.96}
\bibliographystyle{splncs04}
\bibliography{references_lncs}
\endgroup

\end{document}

%% file: abstract.tex
\begin{abstract}
Urban photogrammetric point clouds are commonly processed as local 3D patches, limiting the context available for distinguishing geometrically similar categories. Orthographic projection of the same data yields a co-registered digital orthophoto map (DOM)/RGB-D representation that supports efficient large-context modeling but compresses vertical geometry. Existing projection--point methods mainly fuse these two representations at the feature level while retaining a classifier shared across all samples.
We present the Aerial Content-Aware Sample-Adaptive Classifier (ACASAC), which complements feature fusion with decision-level adaptation. A point-aligned aerial region bridges the full aerial crop and local 3D patch. ACASAC models scene-to-local variation within the projection and the representation discrepancy between aligned 2D and 3D supports, then transfers both relations to refine the local point class representation. The refined representation, scene-level aerial context, and fixed fused-head prior jointly form a classifier for the current sample. ACASAC improves the fixed dual-representation head by 2.1 and 2.7 mIoU points on SensatUrban and UrbanBIS, respectively, and by 2.1--2.3 points across four tested encoder combinations.
\keywords{urban point cloud semantic segmentation \and sample-adaptive classifier \and digital orthophoto map  \and class prototype \and multimodal fusion}
\end{abstract}

%% file: introduction.tex
\section{Introduction}

Semantic segmentation of urban photogrammetric point clouds is fundamental to 3D scene understanding in applications such as autonomous systems, digital twins, and smart-city management. Urban scenes cover substantially larger spatial extents, exhibit strongly non-uniform point densities, and contain semantic categories that are difficult to distinguish from local geometry alone.  Traffic roads, footpaths, and parking areas, for example, may all appear as locally flat surfaces, while their identities depend on surrounding buildings, road networks, vegetation, and object co-occurrence. Reliable urban point-cloud segmentation therefore requires both fine-grained 3D structure and contextual information beyond a sampled point neighborhood. 

A photogrammetric point cloud can be represented either as irregular 3D points or as a regular DOM/RGB-D grid generated by orthographic projection. The point representation preserves local geometry but is typically processed within spatially limited patches, whereas the projected representation enables efficient large-context modeling at the cost of compressed vertical structure. Both representations therefore provide complementary spatial organizations of a common source rather than independent observations.

Projection-based methods use this complementarity to extend contextual reasoning beyond local 3D neighborhoods. BEVSeg~\cite{zouEfficientUrbanscalePoint2021} performs semantic inference in a bird's-eye-view space, while Orth.~\cite{rong3DSemanticSegmentation2023a} predicts from orthographic RGB and elevation images before transferring the resulting labels back to 3D points. 
Dual-branch approaches retain the point representation: APNet~\cite{weiAPNetUrbanlevelScene2023} introduces geometry-aware 2D--3D fusion, and DAPSS~\cite{songDAPSSNovelNetwork2025} incorporates pretrained DOM guidance and adaptive multimodal fusion. These methods improve representation learning, but their final classifier is generally shared across samples.

\begin{figure}[!htb]
    \centering
    \includegraphics[width=1.0\linewidth]{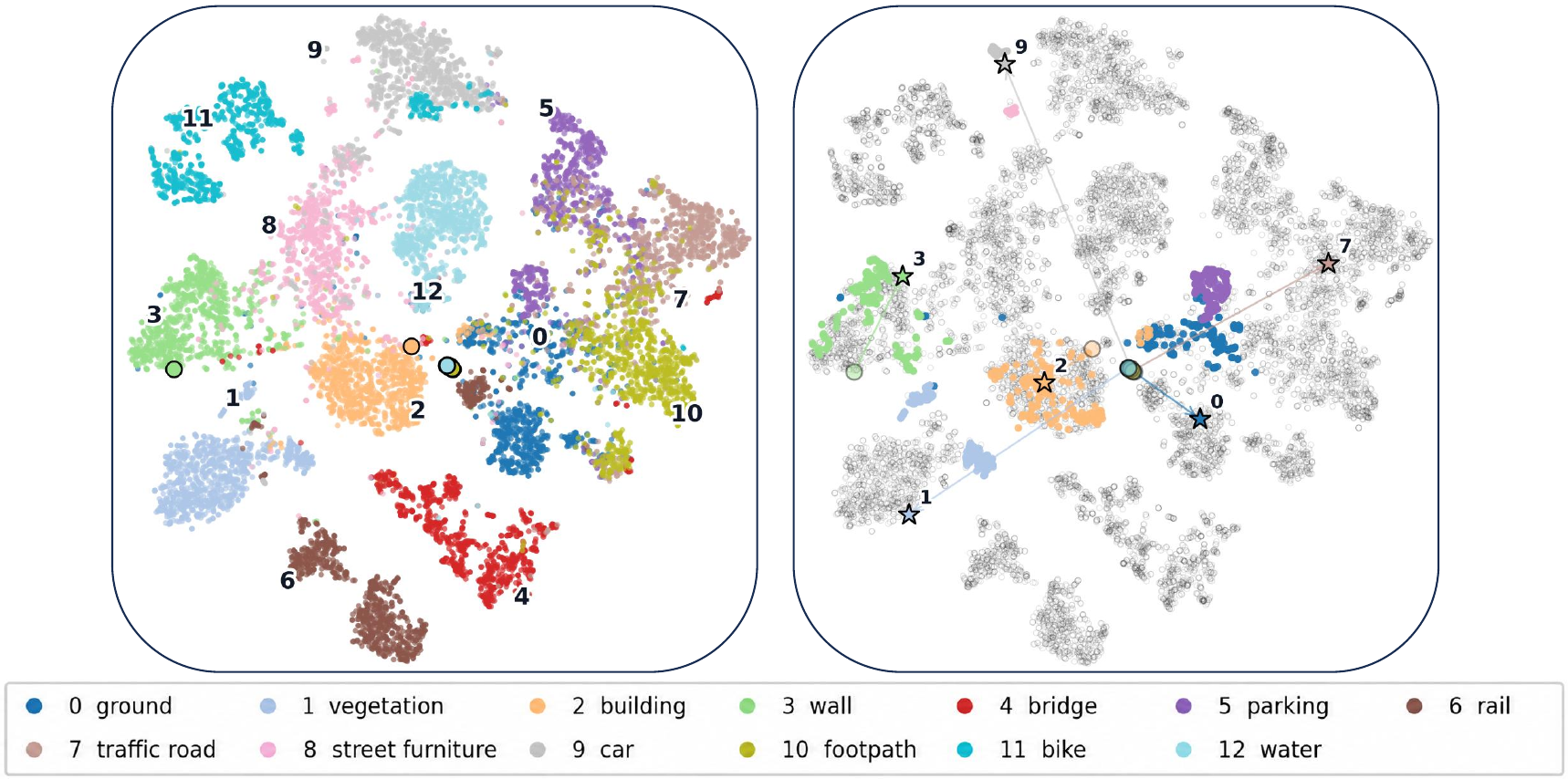}
    \caption{t-SNE visualization of fused point features and classifier prototypes on the SensatUrban test set. The left panel shows the dataset-level feature distribution using 1,000 sampled points per class. The right panel illustrates the sample-specific displacement from fixed classifier prototypes to the adaptive prototypes generated by ACASAC. Small dots denote point features, large circles denote fixed prototypes, and stars denote ACASAC prototypes.}
    \label{fig:tSNE}
\end{figure}

These methods mainly focus on representation-level fusion, while their final classifier is generally shared across all samples. Feature fusion provides each point with additional context, but it does not explicitly adapt the class representations used for prediction. This limitation is relevant to urban scenes, where category presence, frequency, co-occurrence, and spatial layout vary considerably across regions. A globally fixed classifier must accommodate these heterogeneous sample distributions using a single set of decision directions, which may be suboptimal for locally ambiguous categories.
Although sample-adaptive and prototype-based classifiers have been explored in image and point-cloud segmentation~\cite{maSSASegSemanticSpatial2024,tangClassifierEnhancementUsing2025,tianLearningContextawareClassifier2023,zhangCenterguidedClassifierSemantic2025,zhaoNumberAdaptivePrototypeLearning2022,zhouRethinkingSemanticSegmentation2022}, adapting a local point classifier using a broader point-cloud-derived projection has received less attention.

Directly relating the full projected crop to the local point patch is non-trivial because they differ simultaneously in spatial support and representation form. The former covers a broader region through a regular 2D grid, whereas the latter describes a smaller region through irregular 3D samples. We therefore introduce the point-aligned local projected region as an intermediate bridge. The method constructs class prototypes from the full projected crop, the projected region aligned with the sampled points, and the local point features. The relation between the two projected supports characterizes the change from local to broader context, while the relation between the aligned projected and point supports describes how the same local region is represented in 2D and 3D. 
These two relations are transferred to refine the local point class representation, which is combined with scene-level aerial context and the fixed fused-head classifier prior.

Fig.~\ref{fig:tSNE} provides a prototype-based visualization of this motivation. The left panel shows multimodal fusion forms meaningful class clusters at the dataset level, while several categories remain close in the two-dimensional embedding. The right panel illustrates how the adaptive prototypes for one sample are displaced from the fixed classifier prototypes. 
While t-SNE does not directly reflect separability in the original high-dimensional space, these observations are consistent with the sample-dependent feature variation considered here.

The main contributions are summarized as follows:
\begin{itemize}
    \item We formulate urban photogrammetric point-cloud segmentation as a dual-representation decision-adaptation problem. In addition to fusing point-cloud-derived projection features with local 3D features, the proposed method adapts the classifier to the semantic composition of the current projection-crop--point-patch sample.

    \item We introduce a point-aligned local projected representation as a bridge between broader projected context and local 3D observations. It organizes their discrepancy into a cross-scale relation within the projected representation and a locally aligned projection--point relation, which are transferred to correct the local point class representation while retaining the fixed classifier prior.
    
    \item Extensive experiments on SensatUrban and UrbanBIS demonstrate that ACASAC consistently outperforms matched fixed classifiers, CAC, and SSA-Seg under the same multimodal fusion setting. Ablation studies further support the contributions of scene-level aerial context, cross-modal relation transfer, and prototype-guided training.
\end{itemize}

%% file: related_works.tex
\section{Related Work}

\subsection{Point-based Methods}

Point-based methods directly learn geometric representations from irregular point clouds.
PointNet~\cite{qi2017pointnet} introduces permutation-invariant point-wise learning, while PointNet++~\cite{qiPointNetDeepHierarchical2017} extends it with hierarchical local aggregation.
KPConv~\cite{thomas2019kpconv} employs learnable kernel-point convolutions, and Point Transformer~\cite{zhaoPointTransformer2021} models local interactions with self-attention.
For large-scale scenes, RandLA-Net~\cite{huRandLAnetEfficientSemantic2020} greatly improves efficiency through random sampling and lightweight local aggregation, inspiring subsequent urban point-cloud methods~\cite{chen2022multi,dengGANETGlobalAttention2021,huSQNWeaklysupervisedSemantic2022,yooHumanVisionBased2023,yooEyeNetMultiscaleMultidensity2025}.
Nevertheless, point-based networks mainly aggregate information within local 3D neighborhoods, making large-context reasoning relatively expensive.

\subsection{Hybrid Segmentation Methods}

Hybrid methods exploit complementary image context and 3D geometry~\cite{baiTransFusionRobustLiDARcamera2022,liuBEVFusionMultitaskMultisensor2024,zhanSADNetSpaceawareDeepLab2024}. On the one hand, KPRNet~\cite{kochanovKPRNetImprovingProjectionbased2020} maps features from 2D projections back to 3D space, while 3D-MiniNet~\cite{alonso3DMiniNetLearning2D2021} utilizes structured projections for fast processing. Furthermore, Robert et al.~\cite{robertLearningMultiViewAggregation2022} learn to aggregate multi-view image features based on viewing conditions and benchmark different fusion stages.
When applied to aerial urban scenes, BEVSeg~\cite{zouEfficientUrbanscalePoint2021} pioneered the projection of point clouds into a digital orthophoto map (DOM). Subsequently, Orth.~\cite{rong3DSemanticSegmentation2023a} fused DOM and elevation maps using KNN for 2D-to-3D back-projection. Along similar lines, APNet~\cite{weiAPNetUrbanlevelScene2023} achieves geometry-aware multimodal fusion, whereas DAPSS~\cite{songDAPSSNovelNetwork2025} incorporates pretrained DOM priors with adaptive fusion.
Nonetheless, the aforementioned approaches predominantly optimize feature extraction and fusion, leaving the final classification still governed by globally shared parameters.

\subsection{Dynamic Classifiers}

Recent studies~\cite{liangGMMSegGaussianMixture2022,tangClassifierEnhancementUsing2025,zhangCenterguidedClassifierSemantic2025,zhaoFewBigPrototype2025} revisit the classification stage by constructing class representations from contextual information.
ISNet~\cite{jinISNetIntegrateImagelevel2021} introduces class-level contextual representations to enhance pixel features.
CAC~\cite{tianLearningContextawareClassifier2023} generates sample-adaptive semantic prototypes from coarse predictions and contextual features, with a ground-truth-guided classifier providing distillation supervision during training.
SSA-Seg~\cite{maSSASegSemanticSpatial2024} further models both semantic and spatial prototype adaptation: coarse masks guide class-wise semantic and spatial centers, while a training-only teacher constrains adaptation through response-, semantic-, and spatial-domain distillation.
ECAC~\cite{tangClassifierEnhancementUsing2025} additionally employs a memory bank to construct representative global class prototypes.
For 3D point clouds, NAPL~\cite{zhaoNumberAdaptivePrototypeLearning2022} predicts an adaptive number of prototypes for intra-class geometric diversity. In contrast, ACASAC conditions a local point classifier on a broader projection and transfers cross-support and locally aligned projection--point relations.

%% file: method.tex
\section{Method}\label{sec:method}

\subsection{Overview}

\begin{figure}[!htb]
  \centering
  \includegraphics[width=0.94\linewidth]{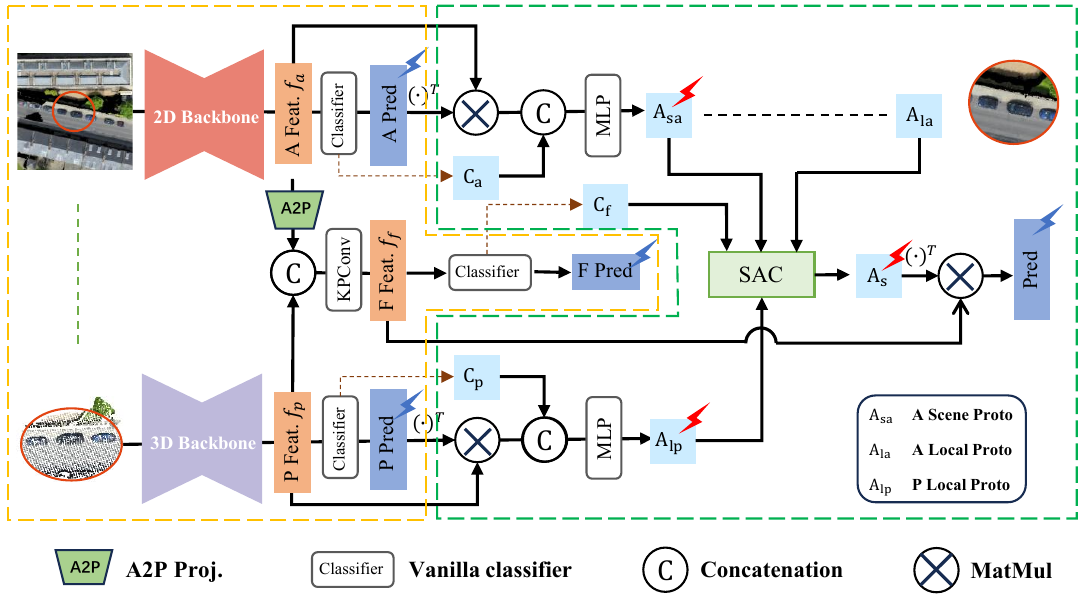}
  \caption{The pipeline of our method. Tensors of the same shape are color-coded identically, with blue and red lightning symbols indicating cross-entropy supervision and knowledge distillation, respectively. The correspondence of A2P is established by the DOM mapping stored during data preparation following~\cite{songDAPSSNovelNetwork2025}.}\label{fig:pipeline}
\end{figure}

Fig.~\ref{fig:pipeline} illustrates the overall pipeline of the proposed method for point cloud semantic segmentation, which integrates dual-modal feature fusion with dynamic classification.

The left dashed box depicts the multi-modal fusion. Given a paired aerial image and point cloud, dual encoders extract image features $\boldsymbol{f}_a\in\mathbb{R}^{HW\times D}$ and point features $\boldsymbol{f}_p\in\mathbb{R}^{N\times D}$. After aerial-to-point projection, KPConv aggregates them into fused point features $\boldsymbol{f}_f\in\mathbb{R}^{N\times D}$. Branch-specific linear classifiers $\boldsymbol{C}_a$, $\boldsymbol{C}_p$, and $\boldsymbol{C}_f\in\mathbb{R}^{C\times D}$ generate coarse predictions for the aerial, point-cloud, and fused branches, respectively.

The right dashed box illustrates the dynamic classification process of ACASAC. Drawing inspiration from CAC~\cite{tianLearningContextawareClassifier2023}, we formulate content-aware prototypes $\boldsymbol{A}\in\mathbb{R}^{C\times D}$ for each modality individually by characterizing and harnessing the latent distributions governing their coarse class-feature spaces. 
Section~3.2 introduces the proposed Sample-Adaptive Classifier (SAC), which models the intra-modal scene-to-local discrepancy within the aerial domain alongside the inter-modal discrepancy across aerial and point-cloud contexts, and derives a scene-conditioned point-cloud prototype $\boldsymbol{A}_{\mathrm{sp}}$. 
Section~3.3 presents the overall training scheme, including the supervision and distillation objectives used to optimize the proposed framework.

\subsection{Sample-Adaptive Classifier}

\begin{figure}[!htb]
  \centering
  \includegraphics[width=0.68\linewidth]{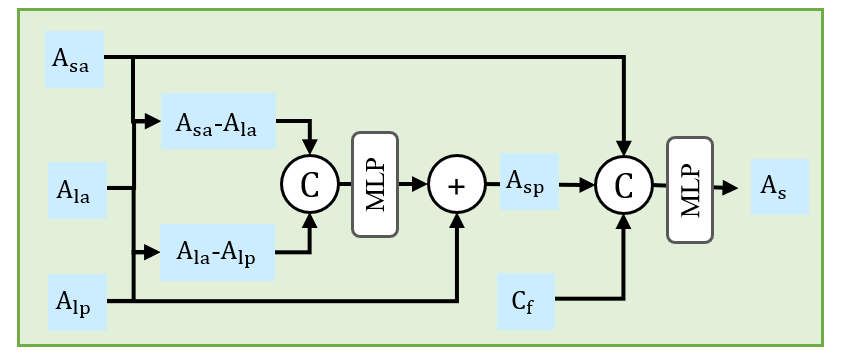}
  \caption{Illustration of the proposed SAC.}\label{fig:sac}
\end{figure}

The core idea of SAC is to transfer two complementary relations: the scene-to-local discrepancy within the aerial modality and the local discrepancy between the aerial and point-cloud modalities. These relations are anchored at the local point-cloud prototypes to predict scene-conditioned point-cloud prototypes.

Following CAC~\cite{tianLearningContextawareClassifier2023}, each unimodal prototype is generated from a branch feature set $\boldsymbol{f}\in\mathbb{R}^{S\times D}$ and its vanilla classifier $\boldsymbol{C}\in\mathbb{R}^{C\times D}$. The coarse probabilities are $\boldsymbol{p}=\sigma(\boldsymbol{f}\boldsymbol{C}^{\top})$, where $\sigma$ applies softmax along the class dimension. They estimate a contextual class center that is combined with the fixed classifier prior:
\begin{equation}
\begin{aligned}
  {(\boldsymbol{C}_{P})}_{c,:}
  &=
  \frac{\sum_{j=1}^{S}p_{j,c}\boldsymbol{f}_{j,:}}
       {\sum_{j=1}^{S}p_{j,c}+\epsilon},\\
  \boldsymbol{A}
  &=\theta(\boldsymbol{C}_{P}\oplus\boldsymbol{C}).
\end{aligned}
\label{equ:Cp}
\end{equation}

Here, $\epsilon=0.01$ stabilizes classes with little support, $\oplus$ denotes concatenation, and the two-layer projector $\theta$ maps each class vector back to width $D$. The scene-level and local aerial prototypes $\boldsymbol{A}_{\mathrm{sa}}$ and $\boldsymbol{A}_{\mathrm{la}}$ share the image-branch classifier and projector, whereas the local point-cloud prototype $\boldsymbol{A}_{\mathrm{lp}}$ uses a separate point-branch classifier and projector. Applying Eq.~\eqref{equ:Cp} to the full aerial sample, its point-aligned local support, and the local point features yields $\boldsymbol{A}_{\mathrm{sa}}$, $\boldsymbol{A}_{\mathrm{la}}$, and $\boldsymbol{A}_{\mathrm{lp}}\in\mathbb{R}^{C\times D}$, respectively.

Although $\boldsymbol{A}_{\mathrm{sa}}$ and $\boldsymbol{A}_{\mathrm{la}}$ are both generated from aerial features, they describe different spatial supports. The former covers the complete aerial sample, whereas the latter is restricted to the point-cloud extent.

As shown in Fig.~\ref{fig:sac}, SAC defines the scene-to-local aerial discrepancy as $\Delta\boldsymbol{A}_{\mathrm{sl}}=\boldsymbol{A}_{\mathrm{sa}}-\boldsymbol{A}_{\mathrm{la}}$ and the locally aligned cross-modal discrepancy as $\Delta\boldsymbol{A}_{\mathrm{l}}=\boldsymbol{A}_{\mathrm{la}}-\boldsymbol{A}_{\mathrm{lp}}$. It then jointly transfers these two relations to the point-cloud domain:
\begin{equation}
\begin{aligned}
  \Delta\boldsymbol{A}_{\mathrm{sp}}
  &=g_r\!\left(\Delta\boldsymbol{A}_{\mathrm{sl}}\oplus\Delta\boldsymbol{A}_{\mathrm{l}}\right),\\
  \boldsymbol{A}_{\mathrm{sp}}
  &=\boldsymbol{A}_{\mathrm{lp}}+\Delta\boldsymbol{A}_{\mathrm{sp}}.
\end{aligned}
\label{equ:relation_transfer}
\end{equation}
The class-wise projector $g_r:\mathbb{R}^{C\times2D}\to\mathbb{R}^{C\times D}$ contains two linear layers. In the final line of Eq.~\eqref{equ:relation_transfer}, $\boldsymbol{A}_{\mathrm{lp}}$ acts as the anchor and the predicted relation residual supplies the scene-level and cross-modal correction, producing the scene-conditioned point-cloud prototype $\boldsymbol{A}_{\mathrm{sp}}$.

The final classifier combines the fixed fused-head prior, the transferred point-cloud prototype, and the scene-level aerial prototype. It then predicts the fused point features by cosine similarity:
\begin{equation}
\begin{aligned}
  \boldsymbol{A}_{s}
  &=\gamma\!\left(\boldsymbol{C}_{f}\oplus\boldsymbol{A}_{\mathrm{sp}}\oplus\boldsymbol{A}_{\mathrm{sa}}\right),\\
  \boldsymbol{Z}^{f}
  &=\eta(\boldsymbol{f}_{f}){\eta(\boldsymbol{A}_{s})}^{\top},\\
  \boldsymbol{P}^{f}
  &=\sigma(\boldsymbol{Z}^{f}).
\end{aligned}
\label{equ:sac_prediction}
\end{equation}
Here, $\gamma:\mathbb{R}^{3D}\to\mathbb{R}^{D}$ is a two-layer class-wise projector, $\eta(\cdot)$ denotes row-wise $\ell_2$ normalization, $\boldsymbol{Z}^{f}\in\mathbb{R}^{N\times C}$ denotes the final logits, and $\boldsymbol{P}^{f}$ denotes the corresponding class probabilities.

\subsection{Training Objective}

As illustrated in Fig.~\ref{fig:pipeline}, the training objective contains three groups that correspond directly to the cumulative configurations in Table~\ref{tab:loss_ablation}: final ACASAC prediction supervision, intermediate representation supervision, and oracle-guided prototype supervision.

\noindent\textbf{Final prediction supervision.}
The final classifier $\boldsymbol{A}_{s}$ produces the logits $\boldsymbol{Z}^{f}$ defined above. Given the point-wise labels $\boldsymbol{Y}^{f}$, the primary segmentation loss is
\begin{equation}
\label{equ:final_ce}
\mathcal{L}_{\mathrm{ce}}
=
\operatorname{CE}(\boldsymbol{Z}^{f},\boldsymbol{Y}^{f}).
\end{equation}

\noindent\textbf{Intermediate representation supervision.}
Branch-specific vanilla classifiers supervise the aerial, point-cloud, and fused representations before the sample-adaptive classifier. Let $\mathcal{B}=\{a,p,f\}$ and $\boldsymbol{Z}_{0}^{b}=\boldsymbol{f}_b\boldsymbol{C}_b^{\top}\in\mathbb{R}^{N_b\times C}$ denote the coarse logits of branch $b$, where $N_a=HW$ and $N_p=N_f=N$. The aerial target $\boldsymbol{Y}^{a}$ contains pixel-wise labels, while $\boldsymbol{Y}^{p}$ and $\boldsymbol{Y}^{f}$ contain point-wise labels. The intermediate representation loss is
\begin{equation}
\label{equ:representation_ce}
\mathcal{L}_{\mathrm{rep}}
=
\sum_{b\in\mathcal{B}}
\operatorname{CE}(\boldsymbol{Z}_{0}^{b},\boldsymbol{Y}^{b}).
\end{equation}

\noindent\textbf{Oracle-guided prototype supervision.}
The adaptive prototypes are estimated from soft coarse predictions and may therefore inherit errors from the early predictions. To provide an ideal training target, we construct a training-only oracle prototype $\boldsymbol{A}^{o}$ from the fused point features $\boldsymbol{f}_{f}$ and their point-wise ground-truth labels $\boldsymbol{Y}^{f}$. Specifically, the one-hot ground-truth assignments replace the coarse probabilities $\boldsymbol{p}$ in Eq.~\eqref{equ:Cp} to compute the oracle class centers $\boldsymbol{C}_{P}^{o}$, which are combined with the fused-head prior and mapped by an oracle-specific projector as $\boldsymbol{A}^{o}=\theta_{o}(\boldsymbol{C}_{P}^{o}\oplus\boldsymbol{C}_{f})$. Because $\theta_{o}$ is independent of the student projectors, the oracle prediction is explicitly optimized by cross-entropy. The resulting $\boldsymbol{A}^{o}$ serves as a ground-truth-conditioned upper-bound target for the dynamic classifier, representing the ideal scene-adaptive prototype under the available labels.

Following CAC~\cite{tianLearningContextawareClassifier2023}, $\boldsymbol{A}^{o}$ then supervises the scene-level aerial prototype $\boldsymbol{A}_{\mathrm{sa}}$, the local point-cloud prototype $\boldsymbol{A}_{\mathrm{lp}}$, and the final dynamic classifier $\boldsymbol{A}_{s}$ through the class-wise KL-divergence paths marked by red lightning in Fig.~\ref{fig:pipeline}. The distillation aligns the class-response distribution induced by each student with that of the oracle. This guidance has two roles: it teaches each student to infer the ideal prototype from its corresponding sample-specific feature distribution, and it aligns the three students within a common class-distribution space. The complete prototype-guidance objective is
\begin{equation}
\mathcal{L}_{\mathrm{proto}}
=
\mathcal{L}_{\mathrm{ce}}^{\mathrm{o}}
+
\mathcal{L}_{\mathrm{KL}}^{\mathrm{sa}}
+
\mathcal{L}_{\mathrm{KL}}^{\mathrm{lp}}
+
\mathcal{L}_{\mathrm{KL}}^{s},
\label{equ:prototype_guidance}
\end{equation}
where $\mathcal{L}_{\mathrm{ce}}^{\mathrm{o}}$ supervises the prediction produced by the oracle-specific projector $\theta_o$, and the other three terms denote the class-wise KL distillation losses for $\boldsymbol{A}_{\mathrm{sa}}$, $\boldsymbol{A}_{\mathrm{lp}}$, and $\boldsymbol{A}_{s}$, respectively, following the design of CAC~\cite{tianLearningContextawareClassifier2023}. As shown in Table~\ref{tab:loss_ablation}, adding $\mathcal{L}_{\mathrm{proto}}$ to $\mathcal{L}_{\mathrm{ce}}+\mathcal{L}_{\mathrm{rep}}$ improves mIoU from 66.67\% to 68.18\%. This cumulative result supports the complete prototype-guidance group but does not isolate the contribution of its four constituent terms. The oracle prototype and its projector are discarded during inference.

The overall training objective is defined as the sum of the three loss groups:
\begin{equation}
\label{equ:total_loss}
\mathcal{L}
=
\mathcal{L}_{\mathrm{ce}}
+
\mathcal{L}_{\mathrm{rep}}
+
\mathcal{L}_{\mathrm{proto}}.
\end{equation}

%% file: experiments.tex
\section{Experiments}

\label{sec:experiments}

\subsection{Experimental Setup}
\textbf{Evaluation Metrics.}
We adopt overall accuracy (OA) and mean intersection over union (mIoU) for evaluation.

\textbf{Datasets.}
SensatUrban~\cite{huSensatUrbanLearningSemantics2022} is a UAV-based urban photogrammetric point-cloud dataset covering 7.64~km$^2$ in three UK cities and 13 semantic classes. We use the official split of 33 training, 4 validation, and 6 test blocks. UrbanBIS~\cite{yangUrbanBISLargescaleBenchmark2023} contains nearly 2.5 billion annotated points from six metropolitan scenes, covering approximately 10.78~km$^2$ and seven classes. We use its official split of 106 training, 39 validation, and 48 test scenes.

\textbf{Implementation Details.}
Following DAPSS~\cite{songDAPSSNovelNetwork2025}, we adopt the same protocol to construct aligned image--point cloud pairs. SensatUrban is voxelized at $0.2~\text{m}$, while UrbanBIS uses its official $0.3~\text{m}$ resolution, No additional data augmentation is applied.. During training, each sample contains $10,000$ locally sampled points and a corresponding $512\times512$ image crop. All models are trained on a single NVIDIA RTX 4090 for 100 epochs with a batch size of 16 using Adam with an initial learning rate of $0.01$, and ExponentialLR with a decay factor of $0.95$. Classifier comparisons use seeds $100$, $1111$, and $1234$.

\subsection{Quantitative And Qualitative Results}

\input{comparison_sota_table.tex}

\begin{figure}[!htb]
    \centering
    \includegraphics[width=1.0\linewidth]{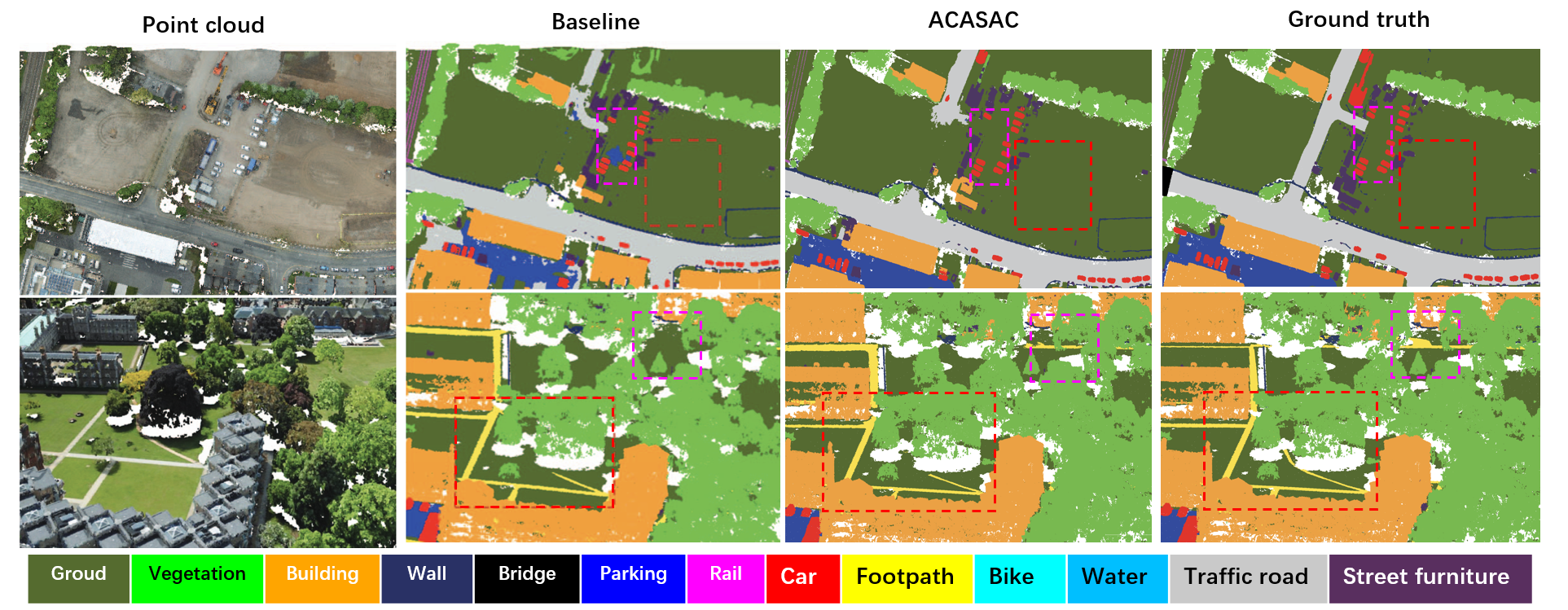}
    \caption{Qualitative comparison on SensatUrban.}
    \label{fig:sensat}
\end{figure}

\begin{figure}[!htb]
    \centering
    \includegraphics[width=1.0\linewidth]{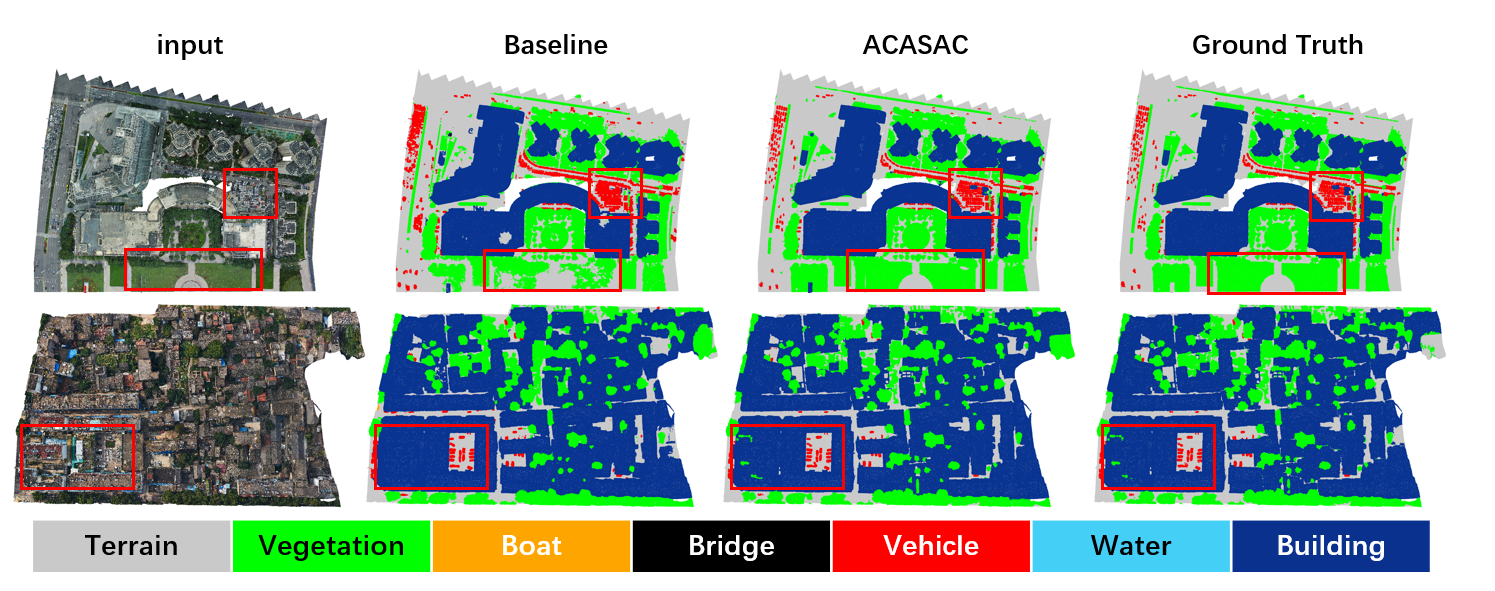}
    \caption{Qualitative comparison on UrbanBIS.}
    \label{fig:urbanbis}
\end{figure}

Figures~\ref{fig:sensat} and~\ref{fig:urbanbis} illustrate how the sample-adaptive head changes local predictions. On SensatUrban, the fixed-head baseline labels the surface beneath a car as parking, whereas ACASAC recovers the correct class by using surrounding context. On UrbanBIS, ACASAC produces cleaner boundaries in noisy residential and parking scenes, especially in mixed-class regions. These examples are consistent with scene-conditioned classification reducing the influence of locally atypical features; they provide qualitative evidence rather than isolating this mechanism.

\input{comparison_sota_classifier.tex}

\input{ablation_on_branch.tex}

\subsection{Ablation Study}

\input{ablation_on_modules.tex}

\input{ablation_on_backbone.tex}

\input{ablation_on_loss.tex}

%% file: comparison_sota_table.tex
\begin{table}
\centering
\caption{Comparison with different methods on the SensatUrban~\cite{huSensatUrbanLearningSemantics2022} and UrbanBIS~\cite{yangUrbanBISLargescaleBenchmark2023} test sets (\%).}
\label{tab:sota}
\small
\setlength{\tabcolsep}{4pt}
\renewcommand{\arraystretch}{1.08}
\begin{tabular*}{\linewidth}{@{\extracolsep{\fill}}llrrrr@{}}
\toprule
 & &  \multicolumn{2}{c}{SensatUrban} & \multicolumn{2}{c}{UrbanBIS} \\
\cmidrule(lr){3-4} \cmidrule(lr){5-6}
Method & Input & OA & mIoU & OA & mIoU \\
\midrule
PointNet++~\cite{qiPointNetDeepHierarchical2017} & PC  & 84.3 & 32.9 & 86.9 & 33.4 \\
RandLA-Net~\cite{huRandLAnetEfficientSemantic2020} & PC   & 89.8 & 52.7 & 92.4 & 61.9 \\
KPConv~\cite{thomas2019kpconv} & PC   & 93.2 & 57.6 & 95.1 & 63.6 \\
PT~\cite{zhaoPointTransformer2021} & PC  & 91.4 & 60.6 & 94.7 & 70.1 \\
PointNAT~\cite{zengPointNATLargescalePoint2024} & PC   & 93.8 & 65.2 & 96.2 & 72.5 \\
BEVSeg~\cite{zouEfficientUrbanscalePoint2021} & RGB   & 91.4 & 61.2 & 91.3 & 62.8 \\
Orth.~\cite{rong3DSemanticSegmentation2023a} & RGBD   & 93.0 & 62.5 & 91.6 & 65.3 \\
APNet~\cite{weiAPNetUrbanlevelScene2023} & PC+RGB    & 94.0 & 65.2  & 94.8 & 72.0 \\
DAPSS~\cite{songDAPSSNovelNetwork2025} & PC+RGBD  & 93.9 & 65.9 & \multicolumn{1}{c}{--} & \multicolumn{1}{c}{--} \\
\midrule
Baseline (Linear)  & PC+RGBD  & 94.2 & 66.1$\pm$0.2 & 95.3 & 72.5$\pm$0.3 \\  
Baseline (Cosine)  & PC+RGBD  & 94.0 & 65.5$\pm$0.2 & 95.1 & 72.0$\pm$0.3 \\ 
\quad +CAC      & PC+RGBD  & 94.3 & 66.9$\pm$0.3 & 95.6 & 73.8$\pm$0.6 \\
\quad +SSA-Seg  & PC+RGBD  & 94.2 & 67.2$\pm$0.4 & 95.4 & 74.1$\pm$0.8 \\  
\quad \textbf{+ACASAC} & PC+RGBD  & \textbf{94.5} & \textbf{68.2}$\pm$0.4 & \textbf{96.3} & \textbf{75.2}$\pm$0.7 \\  
\bottomrule
\end{tabular*}
\end{table}

%% file: comparison_sota_classifier.tex
Table~\ref{tab:sota} compares ACASAC with previously classical methods and with classifier variants evaluated in the shared multimodal framework. All in-house variants share the encoders, fused features, data splits, 100-epoch training schedule, and model-selection protocol. CAC, SSA-Seg, and ACASAC report mean mIoU $\pm$ standard deviation over three random seeds. On SensatUrban, ACASAC achieves 94.5\% OA and 68.2\% mean mIoU, exceeding the fixed linear and cosine heads by 2.1 and 2.7 mIoU points, respectively. On UrbanBIS, its 75.2\% mean mIoU exceeds the two fixed heads by 2.7 and 3.2 points. ACASAC also improves the mean mIoU over CAC and SSA-Seg by 1.3 and 1.0 points on SensatUrban and by 1.4 and 1.1 points on UrbanBIS. Across the three random-seed runs, ACASAC consistently achieves higher mIoU than both adaptive alternatives on the two datasets. The in-house rows therefore compare complete classifier variants under a shared configuration, whereas comparisons with published methods characterize complete-system competitiveness.

%% file: ablation_on_branch.tex
\begin{table}[!htb]
\centering
\caption{Per-class comparison on the SensatUrban~\cite{huSensatUrbanLearningSemantics2022} test set (\%).}
\label{tab:ablation_branch}

\scriptsize
\setlength{\tabcolsep}{2.0pt}

\resizebox{\linewidth}{!}
{%
\begin{tabular}{@{}lrrrrrrrrrrrrrrr@{}}
\toprule
Method
& OA & mIoU & Ground & Veg. & Build. & Wall & Bridge & Park. & Rail & Road & Furn. & Car & Foot. & Bike & Water \\
\midrule

DeepLabV3+~\cite{chenEncoderdecoderAtrousSeparable2018}
& 90.2 & 60.6
& 76.0 & 90.8 & 92.6 & 62.9 & 58.2
& 60.0 & 51.2 & 59.9 & 36.2 & 75.2 & 42.5 & 2.2 & 80.4 \\

RandLA-Net~\cite{huRandLAnetEfficientSemantic2020}
& 91.4 & 52.0
& 82.8 & 91.4 & 90.3 & 58.2 & 22.8
& 48.7 & 24.8 & 62.3 & 30.8 & 73.1 & 39.1 & 0.0 & 51.8 \\

\midrule

Ours(Baseline)
& 94.2 & 66.1
& 87.0 & 98.7 & 96.2 & 77.2 & 58.4
& 62.3 & 55.6 & 61.9 & 48.2 & 84.7 & 45.7 & 0.1 & 83.8 \\

Ours(ACASAC)
& 94.5 & 68.2 
& 87.8 & 98.8 & 96.4 & 77.5 & 64.8
& 65.9 & 60.7 & 63.6 & 49.7 & 84.6 & 49.9 & 2.5 & 84.1 \\

\bottomrule
\end{tabular}%
}
\end{table}

Table~\ref{tab:ablation_branch} shows that multimodal fusion outperforms either unimodal branch, while replacing the fixed head with ACASAC further improves overall performance and benefits 12 of the 13 semantic classes.

%% file: ablation_on_modules.tex
\begin{table}[!htb]
\centering
\caption{Ablation of the prototype information used by ACASAC on SensatUrban (mIoU, \%).}
\label{tab:modules}
\footnotesize
\setlength{\tabcolsep}{7pt}
\renewcommand{\arraystretch}{1.0}

\begin{tabular}{lc}
\toprule
Prototype  & mIoU \\
\midrule
$\gamma(\boldsymbol{C}_{f})$ & 65.71 \\   
$\gamma(\boldsymbol{C}_{f} \oplus \boldsymbol{A}_{\mathrm{la}})$ & 66.88 \\ 
$\gamma(\boldsymbol{C}_{f} \oplus \boldsymbol{A}_{\mathrm{lp}})$ & 66.95 \\  
$\gamma(\boldsymbol{C}_{f} \oplus \boldsymbol{A}_{\mathrm{sa}})$ & 67.31 \\  
$\gamma(\boldsymbol{C}_{f} \oplus \boldsymbol{A}_{\mathrm{lp}} \oplus \boldsymbol{A}_{\mathrm{la}})$ & 67.27 \\
$\gamma(\boldsymbol{C}_{f} \oplus \boldsymbol{A}_{\mathrm{lp}} \oplus \boldsymbol{A}_{\mathrm{sa}})$ & 67.43 \\ 
$\gamma(\boldsymbol{C}_{f} \oplus \boldsymbol{A}_{\mathrm{lp}} \oplus \boldsymbol{A}_{\mathrm{la}} \oplus \boldsymbol{A}_{\mathrm{sa}})$ & 67.76 \\
$\gamma\!\left(\boldsymbol{C}_{f} \oplus \left(\boldsymbol{A}_{\mathrm{lp}} + g_r(\Delta\boldsymbol{A}_{\mathrm{sl}} \oplus \boldsymbol{0})\right)\right)$ & 67.03 \\  
$\gamma\!\left(\boldsymbol{C}_{f} \oplus \left(\boldsymbol{A}_{\mathrm{lp}} + g_r(\boldsymbol{0} \oplus \Delta\boldsymbol{A}_{\mathrm{l}})\right)\right)$ & 67.15 \\
$\gamma\!\left(\boldsymbol{C}_{f} \oplus \left(\boldsymbol{A}_{\mathrm{lp}} + g_r(\Delta\boldsymbol{A}_{\mathrm{sl}} \oplus \Delta\boldsymbol{A}_{\mathrm{l}})\right)\right)$ & 67.64 \\ 
\hline
$\gamma(\boldsymbol{C}_{f} \oplus \boldsymbol{A}_{\mathrm{sp}} \oplus \boldsymbol{A}_{\mathrm{sa}})$ & 68.18 \\ 
\bottomrule
\end{tabular}

\end{table}

Table~\ref{tab:modules} evaluates the prototype information used by
ACASAC. Starting from the fixed classifier prior at 65.71\% mIoU,
introducing the local aerial and local point-cloud prototypes individually
increases performance to 66.88\% and 66.95\%, respectively. Their direct
combination reaches 67.27\%, indicating modest complementarity between the
two locally aligned representations. Among the individual prototypes, the
scene-level aerial prototype achieves the highest result of 67.31\%, which
is consistent with the broader projected support providing a more effective
contextual cue.

Directly combining the local point-cloud and scene-level aerial prototypes
without the local aerial bridge yields 67.43\% mIoU. Adding the local aerial
prototype increases this result to 67.76\%, supporting the usefulness of the
point-aligned aerial representation in connecting the local point and broader
aerial supports. This comparison nevertheless evaluates direct prototype
fusion and does not independently isolate the causal contribution of the
bridge.

Within the relation-transfer variants, using only the scene-to-local or the
locally aligned cross-modal relation obtains 67.03\% and 67.15\% mIoU,
respectively. Jointly modeling both relations raises performance to 67.64\%,
exceeding the two single-relation variants by 0.61 and 0.49 points. The
complete ACASAC configuration further incorporates the scene-level aerial
prototype and achieves the highest tested result of 68.18\%. It outperforms
the relation-transferred prototype alone by 0.54 points and the direct
concatenation of all raw prototypes by 0.42 points. These single-run results
are consistent with the two relations providing complementary information
and with structured relation transfer offering a useful inductive bias over
direct prototype concatenation; they do not establish statistically
significant or independently causal effects.

%% file: ablation_on_backbone.tex
\begin{table}[!htb]
    \centering
    \caption{Effect of multimodal feature encoders on ACASAC on SensatUrban (mIoU, \%).}
    \label{tab:backbone_comparison}
    \footnotesize
    \setlength{\tabcolsep}{5pt}
    \renewcommand{\arraystretch}{1.0}

    \begin{tabular*}{0.82\linewidth}{@{\extracolsep{\fill}}llrrr@{}}
        \toprule
        3D backbone & 2D backbone & Static & ACASAC & $\Delta$ \\
        \midrule

        RandLA & DeepLabV3+ & 66.1 & 68.2 & +2.1 \\
        RandLA & SAM        & 67.5 & 69.7 & +2.2 \\
        PT     & DeepLabV3+ & 66.7 & 68.9 & +2.2 \\
        PT     & SAM        & 68.2 & 70.5 & +2.3 \\

        \bottomrule
    \end{tabular*}
\end{table}

Table~\ref{tab:backbone_comparison} evaluates ACASAC across different point-cloud and image backbone combinations. Although these backbones provide unimodal features of varying quality, ACASAC consistently improves over the matched static head in all tested configurations.

%% file: ablation_on_loss.tex
\begin{table}[!htb]
\centering
\caption{Ablation of the training objectives on SensatUrban (mIoU, \%).}
\label{tab:loss_ablation}
\footnotesize
\setlength{\tabcolsep}{6pt}
\renewcommand{\arraystretch}{1.10}

\begin{tabular}{lr}
\toprule
Training objective & mIoU \\
\midrule
$\mathcal{L} = \mathcal{L}_{\mathrm{ce}}$ & 62.83 \\  
$\mathcal{L} = \mathcal{L}_{\mathrm{ce}} + \mathcal{L}_{\mathrm{rep}}$ & 66.67 \\ 
$\mathcal{L} = \mathcal{L}_{\mathrm{ce}} + \mathcal{L}_{\mathrm{rep}} + \mathcal{L}_{\mathrm{proto}}$ & 68.18 \\
\bottomrule
\end{tabular}

\end{table}

Table~\ref{tab:loss_ablation} shows a cumulative improvement as auxiliary supervision is introduced. Training with only $\mathcal{L}_{\mathrm{ce}}$ yields 62.83\% mIoU. Adding $\mathcal{L}_{\mathrm{rep}}$, which supervises the aerial, point-cloud, and fused intermediate logits, raises performance to 66.67\%, a gain of 3.84 points. Further including $\mathcal{L}_{\mathrm{proto}}$, which combines oracle-guided response distillation and oracle-logit supervision for $\boldsymbol{A}_{\mathrm{sa}}$ and $\boldsymbol{A}_{\mathrm{lp}}$, increases mIoU to 68.18\% by another 1.51 points. The complete objective therefore improves the final-supervision-only setting by 5.35 points. These cumulative results support the contribution of both auxiliary loss groups.